\documentclass{article}

\usepackage{arxiv}

\usepackage[utf8]{inputenc} 
\usepackage[T1]{fontenc}    
\usepackage{hyperref}       
\usepackage{url}            
\usepackage{booktabs}       
\usepackage{amsfonts}       
\usepackage{nicefrac}       
\usepackage{microtype}      
\usepackage{lipsum}
\usepackage{graphicx}
\graphicspath{ {./images/} }

\usepackage{caption}
\usepackage{subcaption}
\usepackage{amsmath}
\usepackage{multirow}

\usepackage{comment}

\usepackage{xcolor}

\newcommand{\rev}[1]{\textcolor{black}{#1}}

\title{UDF-GMA: Uncertainty Disentanglement and Fusion for General Movement Assessment}

\author{
 Zeqi Luo \\
  School of Computing Science\\
  University of Glasgow\\
  Glasgow, Scotland, United Kingdom \\
  \texttt{z.luo.3@research.gla.ac.uk} 
  \And Ali Gooya \\
  School of Computing Science\\
  University of Glasgow\\
  Glasgow, Scotland, United Kingdom \\
  \texttt{Ali.Gooya@glasgow.ac.uk} 
  \And Edmond S. L. Ho\thanks{Corresponding author} \\
  School of Computing Science\\ 
  University of Glasgow\\
  Glasgow, Scotland, United Kingdom \\
  \texttt{Shu-Lim.Ho@glasgow.ac.uk} 
}

\begin{document}
\maketitle
\begin{abstract}
General movement assessment (GMA) is a non-invasive 
tool for the early detection of brain dysfunction through the qualitative assessment of general movements, and the development of automated methods can broaden its application. However, mainstream pose-based automated GMA methods are prone to uncertainty due to limited high-quality data and noisy pose estimation, hindering clinical reliability without reliable uncertainty measures. In this work, we introduce UDF-GMA which explicitly models epistemic uncertainty in model parameters and aleatoric uncertainty from data noise for pose-based automated GMA. UDF-GMA effectively disentangles uncertainties by directly modelling aleatoric uncertainty and estimating epistemic uncertainty through Bayesian approximation. We further propose fusing these uncertainties with the embedded motion representation to enhance class separation. Extensive experiments on the Pmi-GMA benchmark dataset demonstrate the effectiveness and generalisability of the proposed approach in predicting poor repertoire.
\end{abstract}


\section{Introduction}
Prechtl's General Movements Assessment (GMA)~\cite{EINSPIELER199747} has been used as a non-invasive tool for assessing general movements (GMs) of the infant through visual inspection. GMs are spontaneous, whole-body movements in a variable sequence of movements of the arm, leg, neck, and trunk with changing intensity, force, and speed \cite{einspieler2005prechtl}. 
Experienced assessors are able to identify 
atypical movement patterns which may caused by potential brain dysfunction.

GMs can be observed in fetuses as young as 9 weeks postmenstrual age~\cite{de1982emergence}. In infants without neurological dysfunction, GMs follow a consistent pattern from preterm to early post-term age until the end of the second month post-term \cite{EINSPIELER199747}. Before term, these are called fetal or \textit{preterm }GMs; from term age and the first two months post-term, they are known as \textit{writhing movements} (WMs)~\cite{EINSPIELER199747,einspieler2005prechtl}. At the age of 6-9 weeks post-term the pattern of GMs gradually 
transitions into 
\textit{fidgety movements} (FMs)~\cite{EINSPIELER199747}. 
FMs are present up to the end of the first half a year of life, after which intentional and antigravity movements begin to dominate.
GMs are categorised as normal or abnormal based on age \cite{einspieler2005prechtl}. During the preterm and writhing movement period, types of abnormal GMs include \textit{poor repertoire} (PR), \textit{cramped-synchronised} (CS), and \textit{chaotic}. In the fidgety movement period, abnormalities manifest as \textit{abnormal} or \textit{absent}\rev{\textit{/sporadic}} FM patterns. Persistent CS GMs and absent FMs are strong predictors of cerebral palsy (CP)~\cite{prechtl1997early,ferrari2002cramped}.

PR GMs are monotonous and less complex compared to normal GMs~\cite{EINSPIELER199747}. They 
commonly observed in very young infants early in life~\cite{nakajima2006does}, highlighting their role as a typical developmental pattern during early infancy. However, 
PR GMs may evolve into CS GMs or normal GMs over time. This variability underscores the importance of 
assessing early GM patterns, particularly PR GMs, to better predict neurodevelopmental outcomes using GMA~\cite{porro2020early,teschler2023general}. 
Einspieler~\textit{et al.} \cite{einspieler2016general} found that preterm infants with prolonged PR GMs after term exhibited a 5-13 point lower mental developmental index (MDI) or intelligence quotient (IQ) compared to those with normal GMs. 
A study on very preterm infants~\cite{beccaria2012poor} linked PR pattern at 1 month post-term to poorer neurobehavioural outcomes at 2 years of age. These findings highlight the 
importance of identifying 
PR GMs during early infancy to track developmental trajectories and predict potential neurodevelopmental challenges.

Despite its significance, using GMA in practice poses notable challenges.
While experienced GMA assessors can evaluate a recording in 1-3 minutes~\cite{einspieler2005prechtl}, they 
must undergo specialised training and engage in regular practice and recalibration to ensure accuracy. Moreover, effective tracking of developmental progress requires multiple assessments at preterm, term, and post-term stages~\cite{einspieler2005prechtl}. These demands, coupled with the manual nature of the assessment process, have hindered
GMA's widespread clinical adoption. To address these limitations, recent advancements in machine learning 
offer promising solutions
for automating analysis, paving the way for more scalable and accessible GMA applications~\cite{silva2021future}.

Building on these advancements, 
a wide range of automated GMA approaches has been proposed for analysing RGB videos of infants, opening new possibilities for efficient and accurate developmental assessments. Current research focusses on extracting more robust and discriminative motion representation to improve the classification performance, ranging from early work on detecting spatio-temporal movement at the image pixel level using optical flow~\cite{Ihlen:2019,Orlandi:EMBC2018} 
and 2D skeletal poses \cite{McCay:EMBC2019,Chambers:TNSRE2020,McCay:DeepBaby,Sakkos:Access2021,Nguyen-Thai:JBHI2021,Zhu:EMBS2021,Luo:MICCAI2022,Passmore:2024,doroniewicz2020writhing,Pmi-GMA} in more recent approaches, to frequency-based motion features \cite{Stahl:TNSRE2012,Rahmati:TNSRE2016,Zhang:EMBC2022,McCay:TNSRE2022},  which can more effectively de-emphasise high-frequency motion signals likely due to noise. 

Most of the previous work focus on fidgety movements. In particular, Nguyen-Thai \textit{et al.}~\cite{Nguyen-Thai:JBHI2021} proposed a spatial-temporal attention-based model (STAM) that uses Graph Convolutional Network (GCN) with spatial attention to extract clip-level (i.e., short temporal segments) motion representation from skeletal data, then applies attention-based temporal aggregation to obtain video-level classification. Zhu \textit{et al.}~\cite{Zhu:EMBS2021} proposed using channel attention (CA) and a 2D convolutional neural network (CNN) for CP prediction on 3D skeletal data captured using Microsoft Kinect. A recent study by Passmore \textit{et al.}~\cite{Passmore:2024} fine-tuned a pose estimation model for keypoint tracking and labelling to handle infant videos recorded by smartphones. After the skeleton pose extraction, a series of pre-processing approaches are applied to enhance the quality of the pose sequence. A 3-layer 1D CNN is then used for GM classification.

While the methods above formulating the GMA automation as a binary classification problem, WO-GMA~\cite{Luo:MICCAI2022} employed multiple-instance learning (MIL), considered a weakly supervised approach. The authors designed a clip-level pseudo label generation module supervised by the MIL loss, and the video-level label is determined with a top-$k$ strategy. Another stream of work~\cite{McCay:BHI2021,Nguyen-Thai:JBHI2021,Zhu:EMBS2021,Morais:JBHI2023} improves the interpretability of machine learning models by indicating the body parts that contribute more to the GMA prediction outcome.

There are fewer existing work that focus on preterm and writhing GMs. Doroniewicz \textit{et al.}~\cite{doroniewicz2020writhing} propose a writhing movement detection method (WMD) using machine learning to automatically distinguish PR from normal WMs. After pose estimation, they analyse the ellipse circumscribing the movement trajectory to extract features related to limb movement location and scope.
Three algorithms are employed for binary classification: support vector machine (SVM), random forests (RF), and linear discriminant analysis (LDA). In another deep learning-based work, Gong \textit{et al.}~\cite{Pmi-GMA} applied representation learning to CTR-GCN~\cite{CTRGCN} to help the network extract motion representations from 2D pose sequences, emphasising differences between PR and normal preterm GMs.

Although encouraging results were presented in the approaches based on automated skeletal pose-based GMA, two main factors limit the performance of those methods. Firstly, most of the existing methods rely on off-the-shelf pose estimation models such as OpenPose \cite{OpenPose} to extract skeletal pose features. Pose estimation models trained on typical adult body data tend to have lower accuracy when applied to infant videos, and the performance is further affected by video quality and infant self-occlusion. 
This problem can be alleviated by fine-tuning pose estimation models with infant data \cite{Chambers:TNSRE2020,Pmi-GMA,Passmore:2024}, however, a significant amount of manual annotations is needed. 
Secondly, the lack of publicly available data due to privacy and ethical considerations regarding highly sensitive infant videos results in more difficulty training a robust model.

In this work, we propose a new uncertainty-guided deep learning method which tackles the previously stated challenges for automating GMA. Our method models GMA as a binary classification task, using estimated 2D pose sequences to label them as normal or PR GMs. Specifically, the noise and tracking error presented in the estimated skeletal data is modelled as \textit{aleatoric uncertainty} while the uncertainty of the model parameters is represented as \textit{epistemic uncertainty}. By incorporating predictive uncertainty into deep learning, we give a practical formulation to estimate two kinds of uncertainty while performing GMA. Following our formulation, a novel Uncertainty Disentanglement Module (UDM) is designed to provide accurate uncertainty estimation from the motion representation. The epistemic uncertainty is estimated through Monte Carlo Dropout \cite{gal2016dropout}, while the aleatoric uncertainty is directly predicted by the network, training with our novel loss terms. 
To fully exploit the information contained within the estimated uncertainties to improve classification performance, we further propose an Uncertainty-guided motion representation Fusion Module (UFM) to refine the representation embedded by the backbone network. 

We train and validate our method on the Pmi-GMA dataset \cite{Pmi-GMA} that comprises 1120 video clips recorded from 87 preterm infants whose GMs are identified as PR or normal.
Experimental results demonstrate that the proposed model outperformed \rev{the best of recent methods by 5.19\% to 12.09\% across metrics including accuracy, sensitivity, and AUC-ROC}.

The main contributions of this paper can be summarised as:
\begin{itemize}
    \item To our knowledge, this is the first uncertainty-guided deep learning model for automated GMA.
    \item A novel module and loss terms to disentangle aleatoric and epistemic uncertainty during the estimation. 
    \item A new approach to fuse the estimated uncertainties with motion representation for better class separation.
    \item We conducted extensive experiments on the benchmark dataset and compared UDF-GMA with recent automated GMA methods.
\end{itemize}

\section{Preliminaries}
In this section, we will first provide the background on estimating uncertainties in deep neural networks using Bayesian Deep Learning in Section~\ref{section:unc_dnn}. Next, we will introduce the infant movement representation and the motion embedding backbone we used 
(Section~\ref{section:infant_move}). Our data preprocessing pipeline will be explained in Section \ref{section:data_prep}.

\subsection{Uncertainty in Deep Neural Networks} \label{section:unc_dnn}

A deep neural network (DNN) defines a non-linear function: $f_{\omega}:\mathcal{X}\mapsto\mathcal{Y}$ that maps from input space $\mathcal{X}$ to output $\mathcal{Y}$, parametrised by a set of network parameters $\omega$. In a supervised setting such as automated GMA, we have a training dataset $\mathcal{D}=\{\mathbf{x}_i,y_i\}_{i=1}^S \subset\mathcal{X}\times\mathcal{Y}$ containing $S$ pairs of data samples and target. In our dataset, $\mathbf{x}_i$ denotes the pose sequence, and the binary $y_i$ (0 or 1) indicates normal GM or poor repertoire. For an unseen data sample $\mathbf{x}^* \in \mathcal{X}$, a network trained on $\mathcal{D}$ can be used to predict the corresponding target.

We are mainly interested in the \textit{predictive uncertainty} that propagated to the predicted target $y^*$. Network prediction errors can originate from both the model and input data, leading to two types of predictive uncertainty: \textit{epistemic (model) uncertainty} and \textit{aleatoric (data) uncertainty}. Epistemic uncertainty captures the uncertainty about model parameters, which in our case is the model's lack of knowledge due to limited training data available. Aleatoric uncertainty, on the other hand, is related to uncertainty directly arising from the input data, typically caused by information loss such as noise in pose estimation in the automated GMA task. Inspired by \cite{NIPS2017:Uncertainty}, our proposed method is based on Bayesian Deep Learning for modelling the aforementioned types of uncertainty. 

Bayesian neural networks (BNNs) effectively explain uncertainty by replacing deterministic network parameters with distributions. Instead of directly optimising a single set of parameters, BNNs average over all possible sets of parameters.
The learning comes down to Bayesian inference, i.e. computing the posterior $p(\omega|\mathcal{D})$, which describes the uncertainty on the model parameters given a training dataset $\mathcal{D}$ and thus can be understood as an estimation of epistemic uncertainty. The aleatoric uncertainty is then formalised as the uncertainty specified by the probability distribution of the model output $y^*$, given input data sample $\mathbf{x}^*$ and parameters $\omega$. The predictive distribution of $y^*$ is given by

\begin{equation}                p(y^*|\mathbf{x}^*,\mathcal{D})=\int\underbrace{p(y^*|\mathbf{x}^*,\omega)}_{\mathrm{Data}}\underbrace{p(\omega|\mathcal{D})}_{\mathrm{Model}} \mathrm{d}\omega .
\label{eq:pre_dis}
\end{equation}

\subsection{Infant Motion Representation} \label{section:infant_move}

Given an RGB video $V$ with $M$ frames of an infant in the supine position as required by GMA, the 2D skeletal pose in each frame $\mathbf{x}_m=\{x_{m,j}\}^{J}_{j=1} \in \mathbb{R}^{J\times C}$ will be estimated using pose estimation models such as 
HRNet~\cite{HRNet}, and each pose contains the 2D 
coordinates (number of channels $C=2$) 
of $J$ joints. The skeletal motion in the video is then represented by a sequence of skeletal poses $\mathbf{x}=\{\mathbf{x}_m\}^M_{m=1}$.

Pose sequences are high-dimensional. To fully embed the properties of the human skeletal structure and obtain a compact feature representation, GCNs are commonly used. In GCNs for skeletal data modelling, a human skeleton is represented by a 
graph $\mathcal{G}=\{\mathcal{V},\mathcal{E}\}$, where $\mathcal{V}$ is the node set includes $J$ joints in a skeleton and $\mathcal{E}$ is the edge set composed of correlation strength between each two joints which is often formulated as an $J\times J$ adjacency matrix.
With graph representation and stacked layers of graph-temporal convolution operators, GCNs are ideal for spatial-temporal modelling of pose sequences, and achieve SOTA performance in human motion-related tasks such as skeleton-based action recognition \cite{STGCN,CTRGCN}.

In this work, we apply CTR-GCN \cite{CTRGCN} to pose sequences as our motion embedding backbone. This is motivated by: 
(1) The previous SOTA method \cite{Pmi-GMA} on Pmi-GMA dataset is based on CTR-GCN and (2) CTR-GCN is still one of the SOTA models in human action recognition tasks and comparable with more recent GCN models \cite{Lee2023HDGCN,zhou2024blockgcn}. Compared to the very first ST-GCN \cite{STGCN} for pose sequence modelling, the Channel-wise Topology Refinement mechanism in \cite{CTRGCN} refines the graph topology on each channel based on the shared topology among all channels, thus relaxes strict constraints of graph convolutions and improves motion representation.

\subsection{Data Preprocessing} \label{section:data_prep}

\begin{figure}[h]
    \centering
    \includegraphics[width=0.70\linewidth]{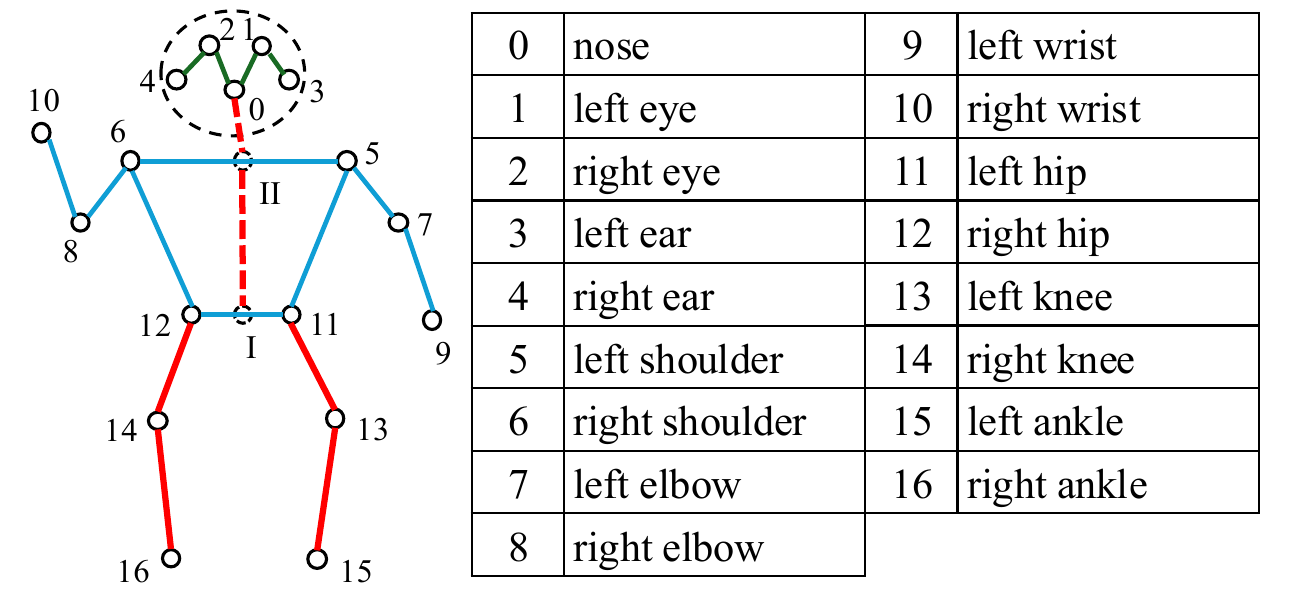}
    \caption{The estimated skeletal pose from a single frame. Joints 0-16 represent body keypoints identified by pose estimation, while positions I and II are calculated for preprocessing.}
    \label{fig:pose}
\end{figure}

The skeletal structure of each pose in the Pmi-GMA dataset 
is shown in Fig. \ref{fig:pose}. We first apply a median filter with a smoothing window of 0.5 seconds to reduce the jitter in the estimated poses. To eliminate the effect caused by camera movement, we shift the skeleton to align the centre of the hips (position I in Fig. \ref{fig:pose}) with the 
origin in each frame and rotate the skeleton to align the trunk (position I to II in Fig. \ref{fig:pose}) with the vertical axis. Since differences in body size may have an impact on the magnitude of body movement, subsequently affecting the classification accuracy, we scale the skeleton to have a normalised height. Specifically, the height of the skeleton is estimated by the sum of the lengths of the following body segments (Fig. \ref{fig:pose}, highlighted in red): lower leg, upper leg, hip centre to neck (trunk) 
and neck 
to nose. 
The authors of Pmi-GMA \cite{Pmi-GMA} report that the fine-tuned pose estimation model accurately identifies all body parts except the ears, which were obscured by the head. In addition, Sakkos \textit{et al.} \cite{Sakkos:Access2021} discovered that using only limb keypoints improves classification performance compared to whole-body keypoints. Therefore, we exclude facial landmarks (keypoints 1 to 4) and perform an ablation study (Section \ref{sec:abl-dp}) on this.


\section{Methodology}
\begin{figure*}[h]
\begin{center}
\includegraphics[width=0.95\textwidth]{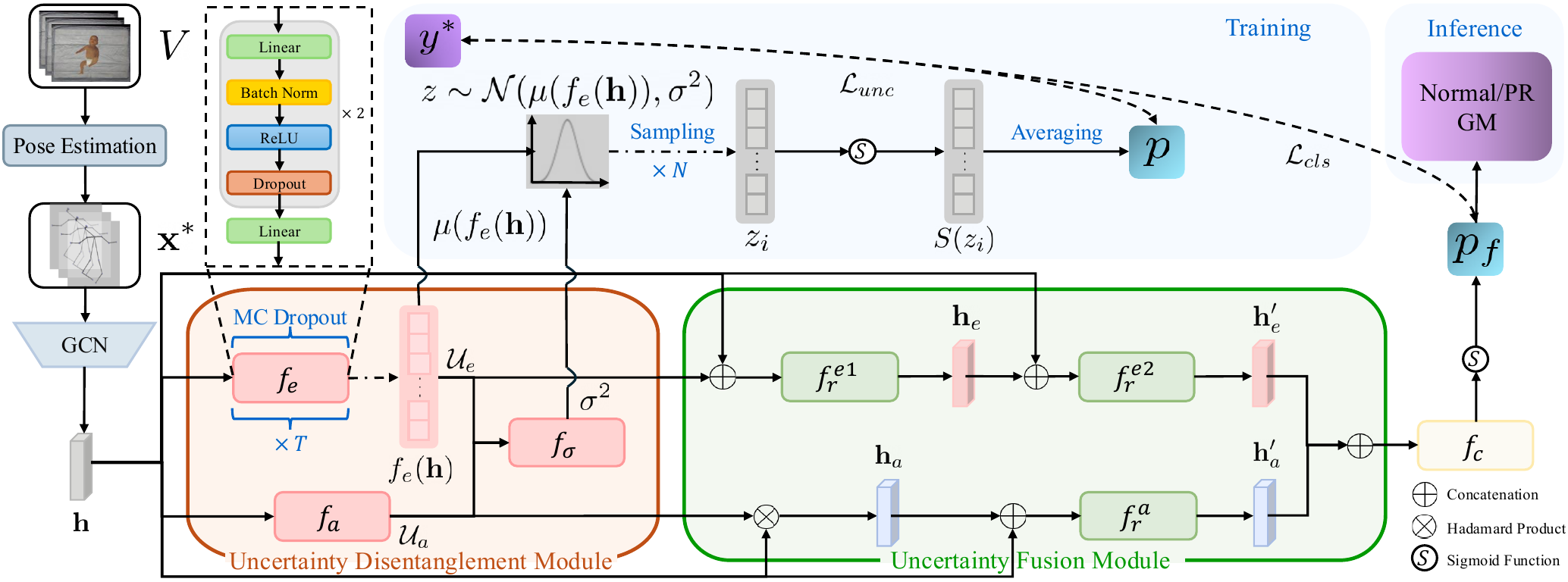}
\caption{Pipeline of UDF-GMA. The proposed methodology focuses on disentangling and estimating the aleatoric (data) uncertainty $\mathcal{U}_a$ and epistemic (model) uncertainty $\mathcal{U}_e$ from the motion embedding $\mathbf{h}$ computed from a GCN backbone. The uncertainties will then be fused with the motion embedding to enhance the performance of binary classification.}
\label{fig:overview}
\end{center}
\end{figure*}

The overview of the methodology is illustrated in Fig.~\ref{fig:overview}. We first give a formulation of the predictive classification probability distribution with two types of uncertainty 
in Section \ref{sec:formulation}. Following the formulation, a dedicated module is designed to disentangle the uncertainties during the estimation (Section \ref{sec:UDM}). Our method models the epistemic uncertainty using Monte Carlo Dropout (Section \ref{sec:formulation}) and the aleatoric uncertainty by a novel loss function (Section \ref{sec:Aleatoric}).
We present the design of the uncertainty fusion module in Section~\ref{sec:fusion}. Finally, the training and inference processes 
are detailed in Section~\ref{sec:t&i}. 

\subsection{Uncertainty Formulation} \label{sec:formulation}
In this work, our first objective is to obtain accurate estimations of epistemic and aleatoric uncertainties, which require the network to disentangle them. Starting from \eqref{eq:pre_dis}, 
the first step is to calculate the posterior $p(\omega|\mathcal{D})$ to represent epistemic uncertainty. However, this is intractable for neural networks with millions of parameters. We leverage Monte Carlo (MC) Dropout \cite{gal2016dropout} which provides an approximating distribution $q_{\theta}(\omega)$ of the posterior, and the model parameters $\omega_{t}\sim q_\theta(\omega)$ are ready to sample. In our binary classification task, given an input $\mathbf{x}^*$, $f_{\omega_t}(\mathbf{x}^*)$ predicts the logit of the positive class (i.e. poor repertoire GM) with a specific set of parameters $\omega_t$ sampled from the dropout distribution. Then the probability of the positive class can be calculated from logit using the Sigmoid function $S(z)=\frac{1}{1 + e^{-z}}$:
\begin{equation}
    \hat{p}_t=S(f_{\omega_t}(\mathbf{x}^*)).
    \label{eq:sigmoid}
\end{equation}
The uncertainty of this probability $\hat{p}$ can then be summarised by the variance, which measures how model is uncertain about this prediction and gives an estimation of $\mathcal{U}_e$:
\begin{equation}
    \mathrm{Var}(\hat{p}) = \frac{1}{T}\sum_{t=1}^T(\hat{p}_t - \bar{\hat{p}})^2
    \label{eq:u_e}
\end{equation}
where $\bar{\hat{p}}$ is the mean of all sampled $\hat{p}_t$.

As we do not directly calculate $p(\omega|\mathcal{D})$, using equation \eqref{eq:pre_dis} to estimate the aleatoric uncertainty is not practical. However, since the network predicts the classification logit, we can reformulate the predictive distribution as follows:
\begin{equation}
    p(y^*=1|\mathbf{x}^*, \mathcal{D}) = \int S(z)p(z|\mathbf{x}^*, \mathcal{D})\mathrm{d}z.
    \label{eq:p}
\end{equation}
When assigning a Gaussian distribution to the logit $z$, the mean is given by the expectation of $f_\omega(\mathbf{x}^*)$ w.r.t. the posterior $p(\omega|\mathcal{D})$, which can be approximated by the mean $\mu(f_\omega(\mathbf{x}^*))$ of $T$ sampled logits:
\begin{equation}
    \mathbb{E}_{p(\omega|\mathcal{D})}[f_\omega(\mathbf{x}^*)] \simeq \frac1T\sum_{t=1}^T f_{\omega_t}(\mathbf{x}^*),
    \label{eq:mu}
\end{equation}
while the variance is accounted for both epistemic and aleatoric uncertainty. We define the total predictive uncertainty as a function of two kinds of uncertainty: $\sigma^2=f_\sigma(\mathcal{U}_a,\mathcal{U}_e)$, then we can formulate the logit distribution:
\begin{equation}
    z \sim \mathcal{N}(\mu(f_\omega(\mathbf{x}^*)), \sigma^2).
    \label{eq:z}
\end{equation}

Using equations \eqref{eq:u_e}, \eqref{eq:p}, and \eqref{eq:z}, we formulate uncertainties in the automated GMA binary classification and offer a practical estimation solution. As aleatoric uncertainty is inherent in the data, we can train the network to directly predict $\mathcal{U}_a$ when performing classification.

\subsection{Uncertainty Disentanglement Module} \label{sec:UDM}

In this section, following our uncertainty formulation, we propose an uncertainty disentanglement module (UDM) to facilitate the estimation of $\mathcal{U}_a$ and $\mathcal{U}_e$.

Given an input pose sequence $\mathbf{x}$, the GCN backbone will first extract the embedded representation: $\mathbf{h}=\mathrm{GCN}(\mathbf{x})$. A common classification network design is to attach a multilayer perceptron (MLP) head to the top of the backbone that takes $\mathbf{h}$ as input and gives the classification logit $z$ as output. In order to disentangle the estimation of two kinds of uncertainty, we split the classification head into two MLPs, $f_{e}$ and $f_{a}$:
\begin{equation}
    z^{\prime}=f_e(\mathbf{h}),\ \mathcal{U}_a=f_a(\mathbf{h}).
    \label{eq:dual_head}
\end{equation}

We denote the output of $f_e$ as $z^{\prime}$ to distinguish it from the predicted logit $z$ in Section \ref{sec:formulation}. When performing MC Dropout, $f_e$ functions as the previously discussed $f_\omega$. The structure of MLP $f_e$ is shown in Fig. \ref{fig:overview}. Each block in the MLP comprises a linear transformation, batch normalisation, activation function, and dropout layer. An additional linear layer is on top of the blocks to produce a scalar output from the high-dimensional hidden representation. Compared to $f_e$, $f_a$ does not have dropout layers to ensure that no additional epistemic uncertainty is introduced to the estimation of the aleatoric uncertainty. In addition, a Softplus activation function $\mathrm{Softplus}(x)=\log(1 + e^x)$ is applied to the output of $f_a$ to produce a legitimate positive output.

After we obtain $\mathcal{U}_e$ and $\mathcal{U}_a$ based on \eqref{eq:u_e} and \eqref{eq:dual_head}, another MLP $f_\sigma$ takes them as input and gives the total predictive uncertainty $\sigma^2$. The structure of $f_\sigma$ is similar to $f_a$, with a Softplus function as the final activation.


Although the network structure with two MLPs is similar to previous work on estimating uncertainty in DNN \cite{NIPS2017:Uncertainty,Ji:ICCV2023-IDM}, our method for defining and estimating uncertainties is improved. \cite{Ji:ICCV2023-IDM} uses two MLPs to predict $\mu$ (mean logit) and $\sigma^2$ (aleatoric uncertainty). They define a probabilistic logit based on these predictions and assess epistemic uncertainty through the entropy of samples from this distribution. However, as the variance of the distribution is defined by aleatoric uncertainty, this measurement mixes both types of uncertainty. In addition, without employing MC Dropout or similar techniques, it cannot capture the uncertainty in predicting $\mu$. Furthermore, previous methods neglect epistemic uncertainty when formulating the logit probability distribution, potentially leading to inaccurate estimations. Our approach addresses the identified shortcomings by clearly defining and disentangling both types of uncertainty, and ensuring the total uncertainty is considered when formulating the predictive distribution.


\subsection{Learning Uncertainty as Loss Attenuation} \label{sec:Aleatoric}

Compared to the calculation of $\mathcal{U}_e$, training the model to predict aleatoric uncertainty is tricky, as we do not have direct supervision for $\sigma^2$ or $\mathcal{U}_a$. However, the classification label $y$ supervises the predicted probability $p$ provided by \eqref{eq:p}, which implicitly includes aleatoric uncertainty in its computation. Generally, the supervision of $p$ uses the binary cross-entropy (BCE) loss $\mathcal{L}_{\mathrm{BCE}}=-\left[y\log(p)+(1-y)\log(1-p)\right]$. Ideally, we want to calculate $p$ through directly integrating the distribution of $z$ given by \eqref{eq:z}. However, there is no analytical solution available. 
As a result, we approximate $p$ through Monte Carlo integration:
\begin{eqnarray}{c}
    z_i = \mu(f_e(\mathbf{h})) + \sigma \epsilon_i,\  \epsilon_i \sim \mathcal{N}(0, 1) \\
    p = \frac{1}{N}\sum_{i=1}^NS(z_i)
\end{eqnarray}
where $N$ is the number of sampling. Note we replace $f_\omega(\mathbf{x}^*)$ in \eqref{eq:z} with $f_e(\mathbf{h})$ defined in Section \ref{sec:UDM}. Then we propose the loss function for learning $\sigma^2$ (and implicitly learning $\mathcal{U}_a$) based on the BCE loss:
\begin{equation}
    \mathcal{L}_{unc} = -\left[y^*\log(p)+(1-y^*)\log(1-p)\right] + \lambda_0(e^{\sigma^2} - 1)
    \label{eq:loss_sigma}
\end{equation}

We observe that with $\mathcal{L}_{unc}$, the learnt uncertainty $\sigma^2$ works as loss attenuation. When the input $\mathbf{x}^*$ and the motion representation $\mathbf{h}$ are noisy, we expect the model to predict an increased aleatoric uncertainty to reflect the high noise level in the data. In this case, the MLP $f_e$ that predicts the logit will struggle to make an accurate prediction, leading to a high BCE loss. Since the high loss punishes the model, it is incentivised to predict a higher $\sigma^2$ to reduce $\mathcal{L}_{unc}$. Consequently, by predicting higher uncertainty to attenuate the impact of noisy data on loss, the model becomes more robust to such data.

At the beginning of training, the prediction can deviate a lot from the ground truth, resulting in a large loss. The model may find 
a ``shortcut" by increasing the uncertainty $\sigma^2$ to make the loss function less sensitive to $\mu(f_e(\mathbf{h}))$ and ignore the data. We further propose the penalty term $e^{\sigma^2} - 1$ in \eqref{eq:loss_sigma} to prevent the uncertainty from taking unreasonably high values. $\lambda_0$ is a trade-off hyper-parameter to temper the strength of the penalty.


\subsection{Uncertainty Fusion Module} \label{sec:fusion}


Up to this point, our proposed model accounts for uncertainties during classification. 
Inspired by the prediction refinement proposed by \cite{Ji:ICCV2023-IDM}, in this section, we propose an uncertainty fusion module that fuses the estimated uncertainty values with the raw motion representation from the backbone to further improve the classification performance.

Previous work \cite{Ji:ICCV2023-IDM} refines the raw embedding by applying a Hadamard product with a weighted sum of model and data uncertainty. This refined representation is then concatenated with the raw embedding to form the final feature. However, we argue that two types of uncertainty can differ in meaning and magnitude, and a simple weighted sum could undermine their representativeness. In addition, the weighted sum introduces extra hyper-parameters. Thus, we propose to fuse the raw embedding $\mathbf{h}$ with $\mathcal{U}_e$ and $\mathcal{U}_a$ separately.

As $\mathcal{U}_e$ is the uncertainty of the model parameters, the physical meaning of the production of $\mathbf{h}$ and $\mathcal{U}_e$ is unclear. Instead, we first design an MLP $f_{r}^{e1}$ to produce the fused representation $\mathbf{h}_{e}$ from the concatenation of $\mathbf{h}$ and $\mathcal{U}_e$. Then, to preserve valuable information in the raw representation and produce a better fusion, another MLP $f_{r}^{e2}$ is applied to the concatenation of $\mathbf{h}$ and $\mathbf{h}_{e}$ along the feature dimension, as:
\begin{equation}
    \mathbf{h}_{e} = f_{r}^{e1}(\mathbf{h} \oplus \mathcal{U}_e), \
    \mathbf{h}_{e}^{\prime} = f_{r}^{e2}(\mathbf{h} \oplus \mathbf{h}_{e}).  
\end{equation}

Unlike epistemic uncertainty, aleatoric uncertainty is a property of the data, so we directly use a Hadamard product of $\mathbf{h}$ and $\mathcal{U}_a$ to produce the fusion $\mathbf{h}_{a}$. A MLP $f_{r}^{a}$ is then applied to the concatenation of raw and fused representation which is similar to computing 
$\mathbf{h}_{e}^{\prime}$:
\begin{equation}
    \mathbf{h}_{a} = \mathbf{h} \odot \mathcal{U}_a,\ 
    \mathbf{h}_{a}^{\prime} = f_{r}^{a}(\mathbf{h} \oplus \mathbf{h}_{a})
\end{equation}

The $\mathbf{h}_{e}^{\prime}$ and $\mathbf{h}_{a}^{\prime}$ are concatenated to form the final fusion representation, which is then sent to the classification network layers $f_c$ for probability prediction:
\begin{equation}
    p_{f} = S(f_{c}(\mathbf{h}_{e}^{\prime} \oplus \mathbf{h}_{a}^{\prime}))
\end{equation}

\subsection{Training and Inference} \label{sec:t&i}
To train our model, we first have a standard BCE loss term to train the whole model for GM classification, where $y$ is the ground truth label and $p_{f}$ is the final prediction:
\begin{equation}
    \mathcal{L}_{cls} = -\left[y\log(p_{f})+(1-y)\log(1-p_{f})\right]
\end{equation}



The network is trained using a total loss function that includes $\mathcal{L}_{unc}$ discussed in Section \ref{sec:Aleatoric} and $\mathcal{L}_{cls}$:
\begin{equation}
    \mathcal{L} = \mathcal{L}_{cls} + \lambda_1\mathcal{L}_{unc}
\end{equation}
where $\lambda_1$ is the hyper-parameter to balance the effect of $\mathcal{L}_{unc}$.

It is important to note that although we apply dropout to all MLPs in our network except $f_a$ and $f_\sigma$, the 
behaviour of these dropout layers during inference is different. In order to estimate the epistemic uncertainty, the dropout layers in $f_e$ still work during inference, while the dropout layers in other MLPs perform as an identity transform.

\section{Experiments}
\subsection{Dataset}

The proposed method is evaluated using the Pmi-GMA dataset~\cite{Pmi-GMA}, which comprises videos collected from 87 newborns, with 23 identified as having PR GMs. The infants were born preterm (gestational age $<$37 weeks) and the videos were recorded within the first week after birth. The dataset includes 1120 segments of 30-second videos from various incubator scenes, with 567 segments classified as PR and 553 as normal. The videos were recorded using stationary smartphones with a resolution of 1280$\times$720 pixels at 30 FPS. All videos were captured from an overhead perspective, with the camera angled in a top-down orientation, while the infant was positioned lying on their back, adhering to the guidelines set by GMA. We perform all experiments using 2D skeletal data extracted at 10 FPS with a fine-tuned HRNet \cite{HRNet} by the authors of \cite{Pmi-GMA} since the raw videos cannot be shared due to privacy concerns. 

Two data partition strategies were proposed in \cite{Pmi-GMA}: \textit{intra} partition, where training, validation, and test sets contain samples from the same infant; and \textit{inter} partition, which ensures that an infant's samples are exclusive to one of these sets. While inter partition is more widely accepted in deep learning research for its ability to test the model's robustness and also a more practical setting in the clinical application, we adopted both strategies due to prior work mainly developed and tested on the intra partition. 

We follow \cite{Pmi-GMA} by allocating 999/121 training/testing samples for the intra partition, and further split 111 samples from the training set as the validation set. For inter partition, we split the training/validation/test set with ratio 0.65/0.15/0.2. Stratified random splitting is used to retain the ratio of PR and normal samples in each set. We perform data shuffling and splitting with different random seeds 5 times, and report the average metrics to evaluate the robustness of the methods. 

Note that no results on the inter partition were reported in \cite{Pmi-GMA}, and the results on the intra partition is non-replicable as their random settings are not provided. We obtained the source code from the authors to run the experiments.


\subsection{Evaluation Metrics}
Following the general setting in classification tasks for medical applications, we report classification accuracy (ACC), sensitivity (SN), and specificity (SP). To enable a detailed comparison with previous automated GMA methods, we also use the area under the curve of the receiver operating characteristics curve (AUC-ROC), as referenced in \cite{Nguyen-Thai:JBHI2021,Luo:MICCAI2022,Passmore:2024}.

\subsection{Implementation Details} \label{section:imple}
\subsubsection{Data Augmentation} Inspired by \cite{Sakkos:Access2021,DataAug}, a series of data augmentation is applied during training to improve the generalisation ability. 
\rev{Specifically, each training sample has an 80\% probability of being augmented. If selected, the following techniques may be applied based on individual probability:}
\begin{itemize}
    \item Mirroring: Flip each frame's pose horizontally by negating the x-pos of its keypoints, with a $50\%$ probability.
    \item Time reversal: Reverse the chronological order of the pose sequence
    , with a $50\%$ probability.
    \item Noise addition: Add Gaussian noise with a mean of 0 and a standard deviation equal to $\frac{1}{3}$ of each channel's standard deviation to the x-pos and y-pos.
    \item Scaling: Adjust the magnitude of the 
    sequence by multiplying all coordinates by a random value in 
    $[0.35,1.65]$.
    \item Magnitude warping: Smoothly adjust the pose magnitude in each frame using different scalars. This is done by generating a random smooth curve varying around $1$ which matches the sample's \rev{duration} 
    and scaling the data by the curve. Applied with a $50\%$ probability.
    \item Time warping: \rev{Distort the time intervals between frames by generating a random smooth curve to simulate variations in time scales. A cumulative version of this random curve is normalised to match the original sequence duration, preserving the temporal structure. The raw data are then interpolated onto these distorted time steps using linear interpolation.}
    Applied with a $50\%$ probability.
\end{itemize}

\subsubsection{Settings} We implemented the proposed methodology and performed experiments with PyTorch 2.4 \cite{paszke2019pytorch} and PyTorch Lightning 2.4 \cite{Falcon_PyTorch_Lightning_2019}. For the motion embedding backbone, we use the same CTR-GCN 
as in \cite{Pmi-GMA}, while the final dropout and linear classification layer have been removed. The proposed modules and the backbone are trained from scratch together, optimised by the SGD optimiser with momentum $0.9$ and weight decay of $0.001$. The base learning rate is set to $0.05$. We train the model for $100$ epochs with $5$ warmup epochs using a multistep scheduler that reduces the base learning rate at the $50$th and $75$th epochs with gamma $0.1$. The batch size is set to $8$. The dropout rate of all MLPs is set to $0.5$. The number of MC dropout sampling is set to $T=100$, which is equal to performing $100$ times stochastic forward propagation of $f_e$. The number of MC integration sampling is set to $N=100$. Weights for loss terms are set to $\lambda_0=\lambda_1=1.0$. We ran the experiments with a NVIDIA GeForce RTX 4090 24GB GPU.

\subsection{Previous Automated GMA Methods}
We compare the proposed method with previous automated GMA approaches \cite{doroniewicz2020writhing,Pmi-GMA} that differentiate PR from normal GMs during preterm and writhing movement period. Additionally, we explore the feasibility of extending prior work on FMs \cite{Luo:MICCAI2022,Passmore:2024,Nguyen-Thai:JBHI2021,Zhu:EMBS2021} to preterm GMs by training and evaluating them on Pmi-GMA. To ensure their optimal performance, we apply the data preprocessing and augmentation from Sections \ref{section:data_prep} and \ref{section:imple} to each method, and tune the hyper-parameters.
For \cite{Luo:MICCAI2022,Passmore:2024,Nguyen-Thai:JBHI2021}, we follow their segment length and stride settings.

\subsection{Comparative Results} \label{sec:results}

The results of our method and the previous ones are summarised in Table~\ref{tab:Pmi-GMA}. Our proposed uncertainty-guided model outperforms previous methods in most metrics. On \textit{intra} partition, our method outperformed all the others in ACC, SP, and AUC-ROC metrics and matched the performance of \cite{Luo:MICCAI2022} in SN. 
Our method further demonstrated more stable performance across all metrics than most of the previous approaches by having a relatively low standard deviation. 
On the more challenging \textit{inter} partition, our method achieved a performance gain of 7.50\% in ACC, 5.19\% in SN and 12.09\% in AUC-ROC over the second-best method. Although our method achieved a lower SP than that reported by \cite{Passmore:2024} and \cite{doroniewicz2020writhing}, we argue that the main application of our method is to serve as a decision support tool for identifying PR. Therefore, a higher sensitivity is more important and thus we optimised our network design and hyper-parameter tuning accordingly. Additionally, our method's standard deviations for ACC, SN, and AUC-ROC are low, while SP's is mid-range compared to other methods. Besides, it is noteworthy that FM-based methods, especially WO-GMA \cite{Luo:MICCAI2022}, achieved results comparable to WM-based methods, highlighting the potential to extend automated GMA methods to different movements.

The results also demonstrated the effectiveness of our proposed uncertainty-guided model. Compared to the base model Gong \textit{et al.}~\cite{Pmi-GMA} which shares the same CTR-GCN backbone as ours on \textit{inter}, our method obtained better results on all 4 metrics: +10.00\% in ACC, +8.74\% in SN, +10.93\% in SP and +16.35\% in AUC-ROC. Furthermore, our method obtained a significantly lower standard deviation (35.13\% to 77.77\%) in all metrics compared to \cite{Pmi-GMA}. This highlights that the model becomes more robust and consistent over different randomly shuffled data splits. We will further analyse the individual uncertainty-guided components in Section~\ref{sec:ablation}.

\begin{table}[!htb]
\centering
\caption{Classification performance on Pmi-GMA dataset. The results are the mean and standard deviation (in brackets) obtained from 5 different runs. All numbers are in \%. We mark the best result in \textbf{bold} and the second-best with an \underline{underline}. \cite{Nguyen-Thai:JBHI2021,Zhu:EMBS2021,Luo:MICCAI2022,Passmore:2024} are originally designed for and validated on fidgety GMs, while \cite{doroniewicz2020writhing,Pmi-GMA} focus on preterm and writhing GMs.}
\label{tab:Pmi-GMA}
\resizebox{\textwidth}{!}{ 
\begin{tabular}{cl|cccc|cccc}
\hline
& \multirow{2}{*}{\textbf{Method}} & \multicolumn{4}{c|}{\textbf{Inter Partition}} & \multicolumn{4}{c}{\textbf{Intra Partition}} \\ \cline{3-10}
                       &  & ACC     & SN       & SP       & AUC-ROC     & ACC     & SN      & SP      & AUC-ROC     \\ \hline
 \multirow{4}{*}{\rotatebox[origin=c]{90}{FM-based}} & CA \cite{Zhu:EMBS2021}   & 54.51(3.39) & 54.74(2.87)  & 54.13(7.88)  & 53.41(6.61) & 75.37(0.96) & 76.55(2.58) & 74.28(0.63) & 81.83(0.72) \\
& STAM \cite{Nguyen-Thai:JBHI2021}  & 58.44(5.71) & 67.44(10.04) & 48.65(15.11) & 57.49(9.03) & 88.26(3.06) & 91.38(4.22) & 85.40(8.36) & 95.94(1.18) \\
& WO-GMA \cite{Luo:MICCAI2022}   & \underline{62.15(4.75)} & \underline{76.54(9.85)}  & 46.57(13.57) & 60.92(8.19) & \underline{93.06(2.89)} & \underline{94.83(3.45)} & \underline{91.43(6.85)} & \underline{98.15(1.31)} \\
& Passmore \textit{et al.} \cite{Passmore:2024} & 58.40(2.65) & 56.54(5.95)  & 60.55(8.39)  & \underline{63.63(3.25)} & 89.75(2.84) & 90.00(3.34) & 89.52(6.40) & 96.41(1.28) \\
\hline
\multirow{4}{*}{\rotatebox[origin=c]{90}{WM-based}} & WMD w/ LDA \cite{doroniewicz2020writhing} & 49.88(8.38) & 52.69(18.09)  & 46.75(3.74)  & 50.15(9.99) & 51.90(1.42) & 46.55(3.45) & 56.83(3.81) & 51.87(1.87) \\
& WMD w/ SVM \cite{doroniewicz2020writhing} & 54.87(4.52) & 44.51(11.17)  & \underline{66.05(5.28)}  & 58.99(5.67) & 84.46(2.05) & 90.00(2.97) & 79.37(1.74) & 91.66(3.07) \\
& WMD w/ RF \cite{doroniewicz2020writhing} & 55.52(8.29) & 43.33(19.31)  & \textbf{68.56(9.95)}  & 56.61(13.18) & 86.61(2.42) & 86.55(2.76) & 86.67(2.77) & 94.32(1.84) \\
& Gong \textit{et al. }\cite{Pmi-GMA} & 59.65(4.68) & 72.99(13.18) & 45.68(19.05) & 59.37(8.77) & 86.28(0.84) & 87.93(2.89) & 82.76(5.38) & 92.98(1.24) \\
\hline
& \textbf{UDF-GMA (Ours)} & \textbf{69.65(1.39)} & \textbf{81.73(8.55)} & 56.61(9.27) & \textbf{75.72(1.95)} & \textbf{93.56(1.10)} & \textbf{94.83(2.44)} & \textbf{92.38(1.56)} & \textbf{98.35(0.54)} \\ \hline
\end{tabular}
}
\end{table}

\subsection{Ablations} \label{sec:ablation}

In this section, we perform comprehensive ablation experiments to demonstrate the effectiveness of our proposed data preprocessing steps, modules, and loss function. Experiments were conducted on the \textit{inter} partition of Pmi-GMA, and we report the mean and standard deviation from five runs.

Since uncertainty lacks a ground truth, we adopt the \textit{uncertainty accuracy} (UA)~\cite{mobiny2021dropconnect} metric which evaluates \textit{epistemic} uncertainty by combining the ground truth label, model prediction, and uncertainty value. It categorises predictions into four types: incorrect-uncertain (\textit{iu}), correct-uncertain (\textit{cu}), correct-certain (\textit{cc}), and incorrect-certain (\textit{ic}). A threshold $\mathcal{U}_t\in[0,1]$ is applied to the normalised uncertainty $\mathcal{U_{\mathrm{norm}}=\frac{\mathcal{U}-\mathcal{U}_{\mathrm{min}}}{\mathcal{U}_{\mathrm{max}}-\mathcal{U}_{\mathrm{min}}}}$ to distinguish between certain ($\mathcal{U_{\mathrm{norm}}} < \mathcal{U}_t$) and uncertain ($\mathcal{U_{\mathrm{norm}}} \geq \mathcal{U}_t$) predictions. A reliable uncertainty estimator should yield correct results when confident (\textit{cc}) and show high uncertainty for likely errors (\textit{iu}). The UA is calculated as the ratio of the desired predictions over all predictions: $\mathrm{UA}({\mathcal{U}_t})=\frac{n_{cc}+n_{iu}}{n_{cc}+n_{iu}+n_{cu}+n_{ic}}$. We report the area under the curve of UA (AUC-UA) with respect to uncertainty threshold $\mathcal{U}_t$ following \cite{mobiny2021dropconnect}, which summarises UA values 
into a scalar.

\subsubsection{Data Preprocessing} \label{sec:abl-dp}
To showcase the effectiveness of proposed data preprocessing steps (Section \ref{section:data_prep}): normalising body size by the sum of different body segments and removing facial landmarks, we perform ablation experiments detailed in Table \ref{tab:abl-dp}. Removing facial landmarks improved all classification metrics. 
For body size normalisation, we tested an alternative method from \cite{Chambers:TNSRE2020,Nguyen-Thai:JBHI2021}, which uses trunk length as a proxy of height. This approach increases SP by 11.09\% but decreases other metrics, notably reducing SN by 12.91\%.

\begin{table}[h]
\centering
\caption{Ablation studies of data preprocessing steps. ``FLs" stands for facial landmarks.}
\label{tab:abl-dp}
\begin{tabular}{l|cccc}
\hline
\multirow{2}{*}{\textbf{Step}} & \multicolumn{4}{c}{\textbf{Classification Performance (\%)}}                                        \\ \cline{2-5} 
                                          & ACC                  & SN                   & SP                    & AUC-ROC              \\ \hline
w/ FLs              & 66.25(2.62)          & 81.40(10.77)         & 50.15(9.27)           & 73.33(8.32)          \\
Trunk Norm                      & 68.45(4.30)          & 68.82(13.27)         & \textbf{67.70(10.74)} & 71.42(5.17)          \\ \hline
\textbf{Ours}                                      & \textbf{69.65(1.39)} & \textbf{81.73(8.55)} & 56.61(9.27)           & \textbf{75.72(1.95)} \\ \hline
\end{tabular}
\end{table}

\subsubsection{Uncertainty Disentanglement and Estimation}
To demonstrate the effectiveness of our uncertainty disentanglement and estimation, we conduct two sets of comparative experiments. First, we evaluate the coarse classification and uncertainty estimation performance of our proposed UEN against previous works \cite{Ji:ICCV2023-IDM, NIPS2017:Uncertainty}. The study by \cite{NIPS2017:Uncertainty} focuses solely on estimating uncertainties without refining classifications. In contrast, \cite{Ji:ICCV2023-IDM} introduces an uncertainty estimation network (UEN) with a prediction refinement module (UPR). For a fair comparison, we exclude the UFM component from our model and the UPR component from \cite{Ji:ICCV2023-IDM}. Second, we substitute the UDM in our model with UEN from~\cite{Ji:ICCV2023-IDM} and the design from~\cite{NIPS2017:Uncertainty}.

In Table \ref{tab:abl}, 
our UDM module 
achieves the highest AUC-UA with the lowest standard deviation, and 
highest ACC and SN among the first comparison group. Although our SP and AUC-ROC are slightly lower compared to \cite{Ji:ICCV2023-IDM}, this is due to 
their direct supervision $\mathcal{L}_\mu$ on coarse predictions, which boosts classification performance. 
The second comparison group also shows the superior performance of UDM in both uncertainty estimation and classification.

\begin{table*}[h]
\centering
\caption{Ablation studies of (1) uncertainty disentanglement and estimation and (2) uncertainty fusion.} 
\label{tab:abl}
\resizebox{\textwidth}{!}{ 
\begin{tabular}{l|cccc|c}
\hline
\multirow{2}{*}{\textbf{Method}} & \multicolumn{4}{c|}{\textbf{Classification Performance (\%)}} & \textbf{Uncertainty Metric (\%)} \\ \cline{2-6} 
                                 & ACC           & SN            & SP            & AUC-ROC        & AUC-UA $\uparrow$                           \\ \hline
UEN \cite{Ji:ICCV2023-IDM}  & 60.46(3.13) & 66.82(24.11)  & 54.26(27.48) & 67.17(8.19) & 44.12(5.56)                  \\
Kendall and Gal \cite{NIPS2017:Uncertainty}  & 57.88(5.31) & 65.84(37.61) & 48.95(32.87) & 63.03(11.44) & 52.19(3.26) \\
UDM (Ours) & 62.71(4.02) & 71.06(11.63) & 53.89(12.44) & 64.64(3.70) & 53.21(1.88) \\
UEN \cite{Ji:ICCV2023-IDM} + UFM (Ours)  & 62.62(4.25) & 67.39(17.05)  & \textbf{57.70(23.77)} & 71.25(6.16) & 45.11(4.91)                  \\
Kendall and Gal \cite{NIPS2017:Uncertainty} + UFM (Ours)  & 59.68(3.98) & 68.61(25.55) & 49.65(20.75) & 63.37(6.25) & 51.56(0.75) \\  \hline

UDM (Ours) + UPR \cite{Ji:ICCV2023-IDM}  & 61.65(4.27) & 72.49(17.89) & 50.24(27.39) & 67.70(8.80) & 53.72(3.59) \\ \hline
\textbf{UDF-GMA, i.e. UDM + UFM (Ours)}  & \textbf{69.65(1.39)} & \textbf{81.73(8.55)} & 56.61(9.27) & \textbf{75.72(1.95)}  & \textbf{58.32(3.30)}                  \\ \hline          
\end{tabular}
}
\end{table*}

\subsubsection{Uncertainty Fusion}
To verify the effectiveness of the proposed uncertainty fusion, 
we compare our method with UDM without any fusion module and UDM with the UPR from \cite{Ji:ICCV2023-IDM} in Table \ref{tab:abl}. The results indicate that adding UDM to the base model~\cite{Pmi-GMA} already improved the performance of ACC, SP, and AUC-ROC. However, incorporating UPR reduced ACC and SP. Replacing UPR with our proposed UFM further enhanced the model, achieving optimal results.

\begin{figure}[h]
    \centering
    \begin{subfigure}[h]{0.25\linewidth}
    \includegraphics[width=1.0\linewidth]{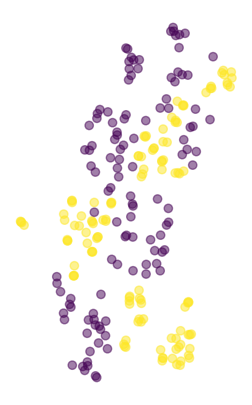}
    \caption{}
    \label{fig:tsne1}
    \end{subfigure}
    \begin{subfigure}[h]{0.25\linewidth}
    \includegraphics[width=1.0\linewidth]{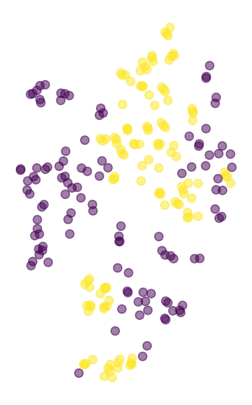}
    \caption{}
    \label{fig:tsne2}
    \end{subfigure}
    \begin{subfigure}[h]{0.25\linewidth}
    \includegraphics[width=1.0\linewidth]{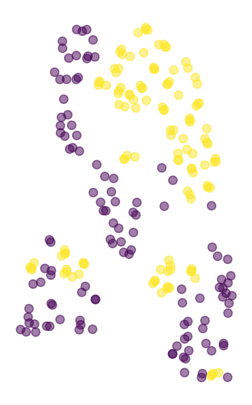}
    \caption{}
    \label{fig:tsne3}
    \end{subfigure}
    \caption{t-SNE visualisation of (a) the raw motion representation given by the backbone, (b) the refined representation by \cite{Ji:ICCV2023-IDM}, and (c) our uncertainty-fused representation of test samples. Purple/yellow dots indicate normal/PR samples.}
    \label{fig:t-sne}
\end{figure}

We employ t-SNE (Fig. \ref{fig:t-sne}) to visualise the representations of test samples from different models. 
Fig.~\ref{fig:tsne3} shows our model achieving the best class separation, with purple and yellow points forming distinct clusters. Fig.~\ref{fig:tsne1} displays moderate separation with some mixing, while Fig.~\ref{fig:tsne2} reveals poor separation as purple and yellow points are heavily intermingled. This suggests that 
UPR~\cite{Ji:ICCV2023-IDM} may compromise the model's representation ability and is unsuitable for automated GMA.

\subsubsection{Loss Function} \label{SSsec:abl-loss}
To analyse the impact and effectiveness of the proposed loss terms, we perform ablation experiments as shown in Table \ref{tab:abl-loss}. Our network outperforms the baseline \cite{Pmi-GMA} even without $\mathcal{L}_{unc}$, demonstrating its effective design. Adding $\mathcal{L}_{unc}$ enhances performance but decreases uncertainty estimation accuracy and consistency across runs due to the absence of a penalty term. Introducing a commonly used $\ell_2$ norm improves uncertainty accuracy but lowers mean classification metrics: ACC, SP, and AUC-ROC. By substituting the penalty term with our proposed exponential term, our model achieves optimal performance across all evaluation metrics. Compared to the commonly used $\ell_2$ norm, our exponential term imposes a stronger penalty on higher $\sigma$, enforcing stricter constraints on excessively high uncertainty values.

In addition to our loss terms, we experimented with the $\mathcal{L}_{\mu}$ loss from \cite{Ji:ICCV2023-IDM}, a BCE loss providing extra supervision to the coarse prediction $\mu(f_e(\mathbf{h}))$. However, incorporating $\mathcal{L}_{\mu}$ negatively impacted model performance. We believe this occurs because our proposed $\mathcal{L}_{unc}$ already implicitly supervises $\mu(f_e(\mathbf{h}))$. Adding $\mathcal{L}_{\mu}$ overly strengthens this supervision, compromising uncertainty estimation and final predictions.

\begin{table*}[h]
\centering
\caption{Ablation studies of different loss terms. $\mathcal{L}_p$ is the penalty term in $\mathcal{L}_{unc}$.}
\label{tab:abl-loss}
\resizebox{\textwidth}{!}{ 
\begin{tabular}{cccc|cccc|c}
\hline
\multicolumn{4}{c|}{\textbf{Loss Term}}                                                   & \multicolumn{4}{c|}{\textbf{Classification Performance (\%)}} & \textbf{Uncertainty Metric (\%)} \\ \hline
$\mathcal{L}_{cls}$ & $\mathcal{L}_{unc}$ & $\mathcal{L}_p$ & $\mathcal{L}_{\mu}$     & ACC           & SN            & SP            & AUC-ROC         & AUC-UA $\uparrow$                           \\ \hline
$\checkmark$        &                       &                     &                       & 61.17(3.43)   & 71.70(12.96)  & 49.87(15.57)  & 68.04(9.26)   &  51.97(1.55)  \\
$\checkmark$        & $\checkmark$     &         &                       &  65.29(4.46)   & 73.36(22.10)  & 56.37(22.26)  & 69.52(11.33)  &   51.87(1.68)   \\
$\checkmark$        & $\checkmark$      & $\frac{1}{2}\sigma^2$        &  &  64.60(4.29)  & 81.40(12.64) & 43.67(16.05) & 65.93(8.68)  &  56.19(3.58)       \\
$\checkmark$        & $\checkmark$      & $e^{\sigma^2}-1$   &  $\checkmark$     & 66.83(2.97)   & 81.50(9.65)   & 50.95(8.57)   & 72.02(4.72)   &  53.22(2.42)    \\ \hline 
$\checkmark$        & $\checkmark$      & $e^{\sigma^2}-1$   &           & \textbf{69.65(1.39)} & \textbf{81.73(8.55)} & \textbf{56.61(9.27)} & \textbf{75.72(1.95)}  & \textbf{58.32(3.30)}  \\ \hline 
\end{tabular}
}
\end{table*}

\subsection{Uncertainty Understanding}
Here, we provide an in-depth analysis of how uncertainties are estimated and their impact on classification performance.

\subsubsection{Epistemic uncertainty} 
\begin{figure}[h] 
\begin{center}
\includegraphics[width=0.5\linewidth]{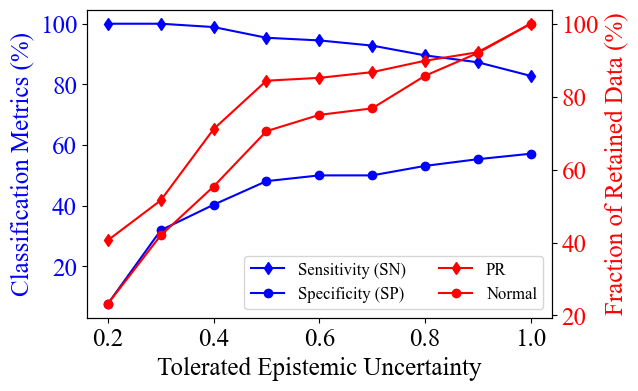}
\caption{The relationship between tolerated epistemic uncertainty with classification metrics (blue) and retention fraction of PR and normal data relative to their original quantities (red). The numbers gained on one of inter partition test sets, and uncertainty values are normalised to between 0 and 1.}
\label{fig:acc_vs_epis}
\end{center}
\end{figure}

arises when a model has insufficient training data, leading to test samples that fall outside the observed data distribution. This is typical in automated GMA tasks due to the scarcity of training samples. 
Understanding decision reliability is crucial, and estimating epistemic uncertainty can help. We conducted an experiment by removing test samples with epistemic uncertainty exceeding certain thresholds and observed changes in metrics and data retention, as shown in Fig. \ref{fig:acc_vs_epis}. Including higher uncertainty samples improved classification SP but reduced SN. This indicates that PR samples with high uncertainty are more likely to be misclassified, while uncertain normal samples have a lower risk of error. When the threshold is below 0.9, the retained proportion of PR samples is higher than normal, suggesting greater model certainty about PR samples. Specifically, at a threshold of 0.2, 40\% of PR samples are retained compared to only 23\% of normal ones.

Our findings indicate that selecting data with lower uncertainty values can ensure a higher sensitivity, which is crucial for medical applications such as GMA. For example, excluding test samples with epistemic uncertainty above 0.4 results in an SN of 98.90\%, while retaining over 70\% PR samples. Practically, setting an uncertainty threshold allows us to identify a certain subset with high sensitivity and flag samples with higher uncertainty for human expert review.

\subsubsection{Aleatoric uncertainty} arises from the inherent randomness or noise in the data, which cannot be eliminated or explained away. In our task, it is intuitive to take into account the quality of the estimated pose when calculating aleatoric uncertainty. 
Unfortunately, in most cases, there is no ground truth skeletal pose to quantify the quality of the estimated pose. As a result, 
confidence values from pose estimation models have been used as additional features~\cite{Ho:CVIU2016} to improve classification performance in previous work. For our method, one potential approach is to add each keypoint's confidence value per frame as an extra input channel, then either embed it separately by a subnetwork or together with the pose using the GCN backbone. However, due to the distribution bias between the training and test data, the pose estimation model can give high confidence to very wrong predictions. Therefore, the use of confidence values in aleatoric uncertainty estimation requires further evaluation.

\begin{figure}[h]
\begin{center}
\includegraphics[width=0.5\linewidth]{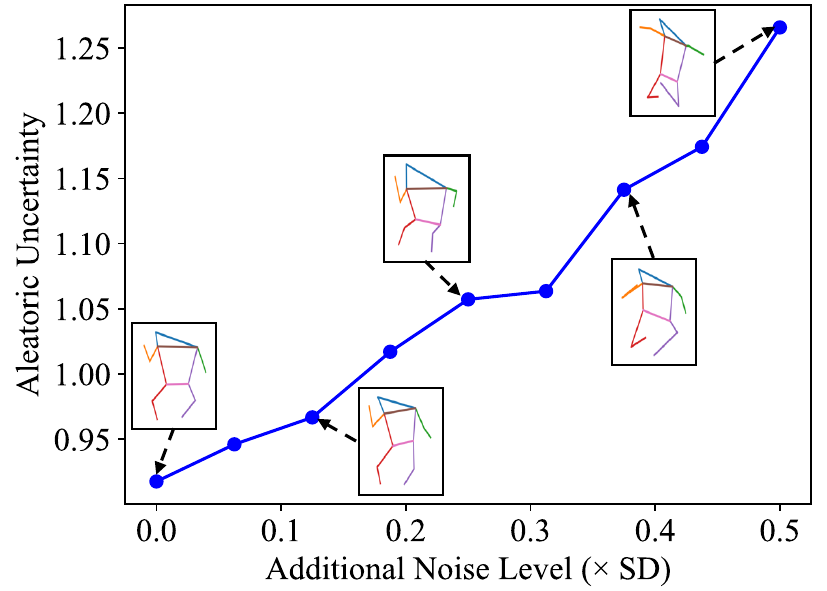}
\caption{Aleatoric uncertainty values estimated from the same pose sequence with different levels of additional noise (from 0.0 to 0.5 Standard Deviation).}
\label{fig:alea_vs_noise}
\end{center}
\end{figure}

Similarly, the Pmi-GMA dataset lacks the ground truth skeletal pose data. 
To understand how aleatoric uncertainty relates to pose noise, we experimented by applying different levels of noise to a randomly selected pose sequence. 
We visualised the estimated aleatoric uncertainty against noise levels. Fig. \ref{fig:alea_vs_noise} shows how the estimated aleatoric uncertainty increases with the degradation of pose quality due to noise. 


\section{Conclusion}


In this work, we present UDF-GMA, the first deep learning method for automated GMA guided by uncertainty. Our approach includes a module specifically designed to disentangle epistemic and aleatoric uncertainties based on our uncertainty formulation. 
Extensive experiments conducted on the benchmark Pmi-GMA dataset demonstrate that our method outperforms recent deep learning-based automated GMA techniques, proving the effectiveness of our modules and loss function design. \rev{In the future, we will include further validation of our approach on other datasets and explore downstream development as a clinical decision support tool for early detection of PR GMs to
track developmental trajectories.}

\section{Acknowledgement}
We would like to thank Professor Cheng Yang for sharing the Pmi-GMA~\cite{Pmi-GMA} dataset and code for reproducing the results, and additional demographic information of the dataset. 

\bibliographystyle{unsrt}  
\bibliography{reference}

\end{document}